\documentclass{article}
\usepackage{ijcai18}
\usepackage{dsfont}
\usepackage{amsmath,amssymb,amsthm}
\usepackage{epsfig}
\usepackage{color}
\usepackage{latexsym}
\usepackage{pdfpages}
\usepackage{csquotes}
\usepackage[tight,footnotesize]{subfigure}
\usepackage{hhline}
\usepackage{bbm}
\usepackage{helvet}
\usepackage{courier}
\usepackage{amssymb,amsmath}
\usepackage{graphicx}
\usepackage{algorithm}
\usepackage{booktabs}
\usepackage{color}
\title{Goldbach's Function Approximation Using Deep Learning}
\author{Avigail Stekel, Merav Chkroun and Amos Azaria\\
 Department of Computer Science, Ariel University, Israel\\
 meravgu@gmail.com, avigail.st@gmail.com, amos.azaria@ariel.ac.il}

\date{\vspace{-5ex}}

\begin{document}
\maketitle
\begin{abstract}
Goldbach conjecture is one of the most famous open mathematical problems. It states that every even number, bigger than two, can be presented as a sum of 2 prime numbers. %
In this work we present a deep learning based model that predicts the number of Goldbach partitions for a given even number. Surprisingly, our model outperforms all state-of-the-art analytically derived estimations for the number of couples, while not requiring prime factorization of the given number. 
We believe that building a model that can accurately predict the number of couples brings us one step closer to solving one of the world most famous open problems. To the best of our knowledge, this is the first attempt to consider machine learning based data-driven methods to approximate open mathematical problems in the field of number theory, and hope that this work will encourage such attempts.
\end{abstract}

\section{Introduction}
On June 1742, the mathematician Christian Goldbach wrote a letter to his friend, Leonard Euler, describing his conjecture that states that every even integer is a sum of two prime numbers \cite{goldbach7letter}. Since then expert mathematicians, students and many others have tried to prove this conjecture or disprove it. Even though more than two hundred and fifty years have passed, the conjecture remains open.
The conjecture can be checked directly for limited sets of numbers. To this date, Goldbach's conjecture has been verified up-to $4\times10^{18}$ \cite{oliveira2014empirical}. During the past centuries, despite no actual proof being found, there has been some important and significant progress related to this conjecture.

\begin{figure*}
\centering
\includegraphics[height=9cm]{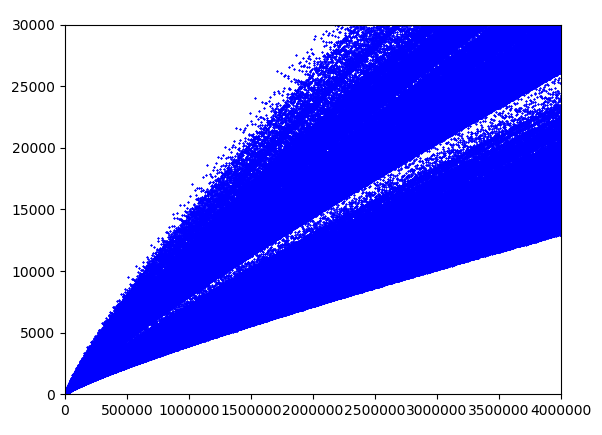} 
\caption{Goldbach function values for all even numbers between $4$ and $4 \times 10^6$. This function is sometimes referred to as Goldbach's comet, due to its shape. }
\label{fig:comet}
\end{figure*}

In this paper, we focus on approximation of the Goldbach's function, denoted by $G(n)$. This function returns the number of Goldbach partitions that a given number has \cite{fliegel1989goldbach}. Rephrasing Goldbach's conjecture in terms of Goldbach's function would state that the value of Goldbach's function (for all even numbers greater than 4) is greater than or equal to 1.
For example, $G(100) = 6$, because $100 = 3 + 97 = 11 + 89 = 17 + 83 = 29 + 71 = 41 + 59 = 47 + 53$.
See Figure \ref{fig:functionTill200} for an illustration of Goldbach's function on the first $100$ even numbers.
The plot of the Goldbach function has a form of a comet and is consequently called ``Goldbach's comet" \cite{fliegel1989goldbach} (See Figure \ref{fig:comet} for Goldbach's function values for all even numbers between $4$ and $4 \times 10^6$. 

Several works have suggested different approximations for Goldbach's function which they have derived analytically. Unfortunately, some of these approximations are very far from the actual values taken by Goldbach's function, while others require prime factorization (prime decomposition) which is  believed to be an intractable operation on large numbers.

In this paper we suggest a different approach to approximating Goldbach's function, we propose using a deep learning approach. It may seem that deep learning is not a suitable approach for this type of problems, as the input to the approximation function is only a single number, and deep learning has shown success in situations in which the input is composed of a large vector or a matrix. We therefore propose a simple, yet powerful concept of translating the number into different bases. Surprisingly our approach outperforms current state-of-the-art approximations of Goldbach's function, resulting in an error rate of only $3.0\%$. Furthermore, our method does not require prime factorization of the number, which is intractable for large numbers.
We believe that our model may shed light on the behavior of Goldbach's function and may bring us one step closer to proving or disproving Goldbach's conjecture. Furthermore, introducing deep learning to the field, may assist in proving or disproving other similar open mathematical problems.


\section{Background}
Prime and natural numbers have always aroused mathematicians' interest. In 1900 Hillbert made his famous speech at the 2nd International Congress of Mathematics held in Paris, saying there are $23$ unsolved problems for mathematicians of the 20th century \cite{wang2002goldbach}. One of those math problems was Goldbach's conjecture.

\subsection{Approximations for Goldbach's Function}
Goldbach's Conjecture is divided into two conjectures:
\begin{enumerate}
\item The `weak' Goldbach's conjecture states that `Every odd number greater than 5 can be expressed as a sum of three primes'. For example, $11$ is the sum of $3$, $3$ and $5$. $21$ is the sum of $2$,  $2$ and $17$. The weak conjecture was finally proved in 2013 \cite{helfgott2013ternary}.
\item The `strong' Goldbach's conjectures which states that `Every \emph{even} integer greater than $2$ is a sum of two primes'. The number $6$ for example, can be presented with only one pair of prime numbers , $3+3$ . However, when examining even numbers greater than $12$ , there are apparently, at least two pairs of prime numbers that sum to each even number, for example, $14 =7+7$ and $14=3+11$. One might assume that the greater the even number, the more different pairs it has, yet by observing different even numbers this assumption turns out to not always hold. For example, while $34$ and $36$ have $4$ Goldbach partitions each, $38$ has only $2$ Goldbach partitions as is shown in Figure \ref{fig:functionTill200}. This conjecture remains open until this date.

\begin{figure*}
\centering
\includegraphics[height=9cm]{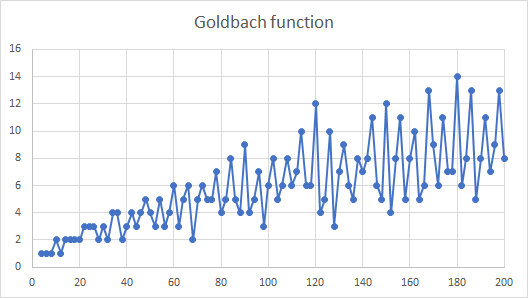} 
\caption{The number of Goldbach partitions of a couple of even numbers}
\label{fig:functionTill200}
\end{figure*}
\end{enumerate}

In this paper we focus on Goldbach's function which provides the number of Godlbach's partitions an even number has.
More formally, Let $n\in \mathbb{N}$, the Goldbach's function is given by: 
\begin{equation}
    G(n)= \sum_{\{p,q\}\in \mathbb{P}\times \mathbb{P} \: \wedge \: p\leq q} {\mathbbm{1}\{p+q=n\}}
\end{equation}
where, $\mathbb{P}$ is the set of all prime numbers, and $\mathbbm{1}$ is the indicator function that returns $1$ if the expression is \emph{true} and $0$ otherwise.

Over the years there have been several attempts to find an analytic approximation to Goldbach's function.
Hardy and Littlewood \cite{hardy1922some} proposed the following approximation: 

\begin{equation}
    G_1(n)= 2 \cdot C_2 \frac{n}{(ln(n))^2}\prod_ {p|n}(\frac{p-1}{p-2})
\end{equation}
where $C_2$ is their twin prime constant: 
\begin{equation}
   C_2 = \prod_{p\ge3}\left(1-\frac{1}{(p-1)^2}\right) \cong 0.6601618158
\end{equation}
$n$ denotes an even number, and $p$ denotes all the prime factors of $n$. While this function was originally proposed as an upper-bound, it is widely used as an approximation. 
Baker, suggested multiplying $G_1(n)$ by $\frac{3}{5}$ to yield a better approximation \cite{baker2007excel} (we will refer to Baker's approximation as $G_2(n)$).

Note that this approximation requires factorizing $n$, which is assumed to be a hard problem. Currently, best known prime factorization algorithm (GNFS) \cite{buhler1993factoring} runs in time complexity of:
\begin{align*}
    O\Bigg(exp\bigg(\Big(\sqrt[3]{\frac{64}{9}} + o(1)\Big)\sqrt[3]{log(n)}\sqrt[3]{(log log(n))^2}\bigg)\Bigg)
\end{align*}
where $n$ is the number being factored. Note that the input size is considered $log (n)$, since the number of digits required to represent $n$ is $log(n)$.

To overcome this prime factorization requirement, the following approximation was proposed \cite{provatidis2013rule}: 
\begin{equation}
    G_3(n)= \frac{n}{(ln(n))^2}
\end{equation}
This approximation is derived from Gauss' approximation provided in 1793 for the probability of a number being prime. According to Gauss, this probability is given by:
\begin{equation*}
  f(m) = \frac{m}{ln(m)}
\end{equation*}
Therefore, for an even number $n$ the following may be used as an approximation for its number of Goldbach partitions:
\begin{equation*}
  	\sum_{m=3}^{n/2}{\frac{m}{ln(m)}\cdot\frac{n-m}{ln(n-m)}}\approx \frac{n}{2ln(n)^2}
\end{equation*}
Note that $G_3$ is monotone, and thus cannot capture the phenomenon that larger numbers may sometimes have less Goldbach partitions than smaller numbers.
The following approximation, which is also monotone, was proposed by Markakis et al. in \cite{markakis2012some}:
\begin{equation}
    G_4(n)= \frac{n}{(ln(n/2))^2} 
\end{equation}


\subsection{Related Work}
In addition to attempts for finding an approximation to Goldbach's function, there have been several attempts to finding upper and lower-bound to it, that is, a function that limits the number of Goldbach partitions from above or below. The $G_1(n)$ function proposed by Hardy and Littlewood \cite{hardy1922some} was originally suggested as an upper bound.
One proposed lower-bound function provided by Provatidis et al. \cite{provatidis2013rule} is:
\begin{equation}
    2/3*G_1(n)
\end{equation} 
This lower-bound was derived analytically, and it is shown that as $n$ grows, the probability of it having less Goldbach partitions than the lower bound approaches 0. However, proving this lower-bound as a strict lower-bound, would imply the proof of Goldbach's conjecture, since this lower-bound is at least 1 for every even number.

Montgomery and Vaughan define another function related to Goldbach's conjecture, capturing non-Goldbach numbers, that is, numbers that cannot be written as a sum of two prime numbers \cite{montgomery1975exceptional}. Montgomery and Vaughan's function, $E(n)$, denotes all even numbers smaller than $n$ that are not a Goldbach number. Montgomery and Vaughan prove that there exists an absolute constant $\delta>0$ such that 
\begin{equation}
    E(N) \ll N^{1-\delta}.
\end{equation}

There are several fields lying in the intersect of artificial intelligence and mathematical problems. 
Automated theorem proving \cite{bibel2013automated} is a field in which machines use various artificial intelligence based methods, such as heuristic search, in an attempt to find a proof for a given conjecture. In 1956, Newell and Simon developed the ``Logic Theorist'' \cite{newell1956logic}. The Logic Theorist was based on heuristic search and successfully proved $38$ of the $52$ theorems that appear in the second Chapter of Principia Mathematica \cite{whitehead1912principia}.

The `Automated Mathematician' (AM for short) was created by Douglas Lenat in Lisp.\cite{lenat1977automated}. AM used heuristic search to find interesting properties in mathematics. AM defined $250$ various heuristics and tried to infer different mathematical properties by applying these heuristics. AM discovered the concept of natural numbers, prime numbers, it conjectured (without proof) the unique prime factorization theory and defined the concept of Goldbach partitions. Unfortunately, AM was not able to discover any ``new to mankind'' mathematics, and it turned out to be very hard for it to discover new heuristics. One of the statements Lenat's AM produced was the Goldbach conjecture \cite{larson2005survey}. AM was more about finding interesting problems than solving them. An improved system named EURISKO was later developed by Lenat, with an attempt to learn these heuristics by its own \cite{lenat1983eurisko,lenat1984and}.

Colton et al. \cite{colton2000automatic} developed an artificial intelligent system for identifying  mathematical concepts, such as, types of graphs, types of groups and types of numbers. For example, their method can identify that a sequence such as 1, 4, 9, 16 etc. is a sequence of squared numbers. They state that the state-of-the-art at their time for identifying these concepts was just a data-base.

\section{Deep Learning Based Goldbach's Function Approximation} 
In this section we present a deep learning based model to approximate Goldbach's function values.

\subsection{Data Composition}
In order to train and evaluate the different methods, we composed a data-set consisting of the number of Goldbach partitions that all even numbers from $4$ to $4\times 10^6$ have. To that end, we first computed all prime numbers at that range, and stored them as a list and as a hashmap. For each even number, $n$, we iterated on all prime numbers (using the list of all primes) that are smaller than or equal to $\frac{n}{2}$. For each of these prime numbers, $p$ we test (using the hashmap) whether $n-p$ is a prime number itself. If so, we increment $n$'s counter by one. 

We shuffled the data and split it into a train set, ($80$\% of the data; $16\times 10^5$ numbers), a validation-set, ($10$\% of the data; $2\times 10^5$ numbers), and the remaining $10\%$ was reserved for the test-set ($2\times 10^5$ numbers).   


\subsection{Model Features}
From each number we extracted $42$ features. We converted every number to its binary representation, ternary representation (base $3$), quinary representation (base $5$) and septenary representation (base $7$). The time complexity of computing these base representations for a number $n$ is $O(log(n))$. In practice we computed those representations when composing the data, so we simply incriminated the representation of the previous number by $2$ for all bases.
We truncated these representations and used the $10$ least significant digits for each representation. The intuition behind using these different representations lies in the fact that these transformation are computationally cheap to extract and that they might allow the model to retrieve underlying information on the number. The first $4$ prime numbers ($2,3,5,7$) were selected as the bases.
In addition to the representations in the different bases, we added the number itself (normalized), and the logarithm of that number.

\subsection{Model Architecture}
We used a fully connected neural network as our model. We set the number of neurons to $200$ on each hidden layer. We used Adam optimizer \cite{kingma2014adam}, with a learning rate of $0.001$. We used a mini-batch size of $1024$ and trained the model for approximately $200$ epochs on the data. We used early stopping \cite{prechelt1998automatic}, that is, we evaluated the validation set every epoch and saved the variables which obtained the lowest validation error.
We varied the number of hidden layers, starting at a simple linear regression model (with no hidden layers), a model with 3 hidden layers, 5 hidden layers, and 7 hidden layers. Each of these models was trained on the training data and their performance was evaluated on the validation set. See Table \ref{tbl:depth} for a summary of the validation results. As can be seen in the table, the model with 5 hidden layers performed best on the validation set, and was therefore chosen as the model for our further analysis. For a given number $n$, the time complexity of generating the features and evaluating our model is $O(log(n))$, which is the best time complexity one could expect from an algorithm that reads the entire input (which requires $O(log(n))$ digits to represent).


\begin{table}[h!]
\centering
 \begin{tabular}{|c | c c |} 
 \hline
  & Train MSE & Validation MSE\\ [0.5ex] 
\hhline{|=|==|}
 Linear regression & 960,400 & 1,016,064  \\ 
 \hline
 3 hidden layers  & 92,933 & 107,223 \\
 \hline
 5 hidden layers & 89,764 & \bf{103,457}  \\
 \hline
 7 hidden layers & 88,446 & 105,903  \\
 \hline
\end{tabular}
\caption{Train and validation mean squared error (MSE) according to the number of hidden layers. We select the model with the lowest validation error (5-hidden layers). }
\label{tbl:depth}
\end{table}

\begin{table*}[h!]
\centering
 \begin{tabular}{|c | c c c |} 
 \hline
  & MSE on test & RMSE on test & Error rate\\ [0.5ex] 
\hhline{|=|===|}
 $G_1$ \cite{hardy1922some}* & 89,059,989 & 9437.1 & 87.6\% \\ 
 \hline
 $G_2$ \cite{baker2007excel}* & 221,437 & 470.57 & 4.4\% \\
 \hline
 $G_3$ \cite{provatidis2013rule} & 24,902,559 & 4990.3 & 46.3\%  \\
 \hline
  $G_4$ \cite{markakis2012some} & 22,517,117& 4745.2 & 44.0\% \\
  \hline
 Deep-learning based method & {\bf 105,100} & {\bf 324} & {\bf 3.0\%}  \\
 \hline
\end{tabular}
\caption{Approximation error of the deep learning based method in comparison to the state-of-the-art approximations. Asterisk (*) denotes models that require prime factorization of the given number.}
\label{tbl:results}
\end{table*}

\subsection{Results}
\begin{figure*}
\centering
\includegraphics[height=10cm]{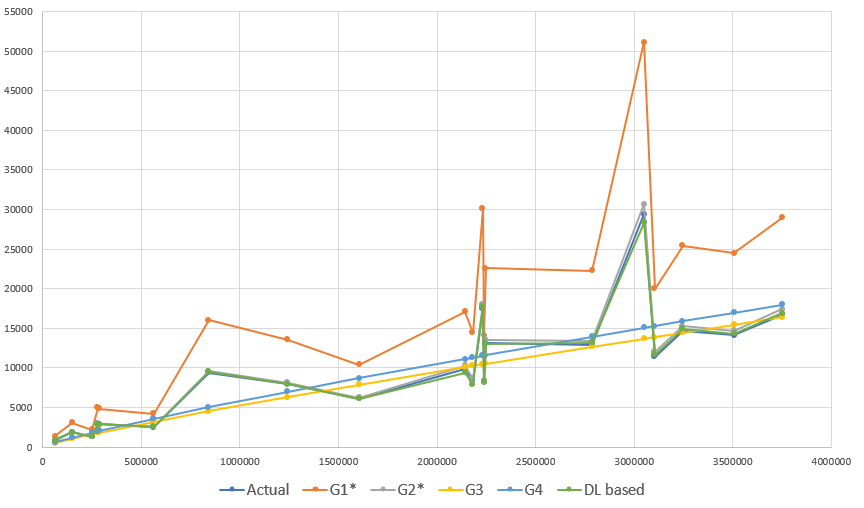} 
\caption{Prediction of the different methods on 20 randomly picked numbers from the test-set. The asterisk (*) denotes models that require prime factorization of the given number.}
\label{fig:resultExample}
\end{figure*}

Table \ref{tbl:results} presents the performance of our model in comparison to the formulas that appear in the literature, in terms of mean squared error (MSE), root mean squared error (RMSE) and the error rate in comparison to the number of actual pairs (that is, the RMSE divided by the mean of the number of Goldbach partitions each number in the test-set has). As can be seen in the table, our model outperformed all previous approximation attempts, achieving a new state-of-the-art approximation model. Furthermore, our model does not require factorizing the given number.
Figure \ref{fig:resultExample} compares the approximation of the different methods on 20 randomly sampled numbers from the test-set.
As illustrated in the figure, our approach achieves the best fit to the actual points.
While $G_3$ and $G_4$ follow the average value of Goldbach's function, they do not follow the ups and downs of it. $G_1$ does follow the ups and downs of Goldbach's function but keeps a gap all long the plot. While this gap is corrected nicely by $G_2$, $G_2$ (as well as $G_1$) requires prime factorization to be computed. 


Using our trained model, we tried to articulate what a number violating Goldbach's conjecture may look like. We used a hill climbing search method on the base representations of the input features to the model. We set the number itself to $10^6$ and its log value accordingly. Iteratively, we traversed each of the digits of each of the base representations, searching for the digit value that minimizes our model's prediction. We repeated this process until no digit was changed.
Table \ref{tbl:magicalNum} presents the base representations of a hypothetical number found by the search method. According to our model, this number violates Goldbach's conjecture, with a prediction of -192,886 Goldbach partitions (note the negative value). This number is a factor of 14, has a remainder of 2 when divided by 3 and a remainder of 4 when divided by 5.
We performed a search on numbers satisfying base 7 representation, that is, numbers of the form $m\times 7^{10} + 6\times7^9 + 7^8 + 6\times 7^2 + 4 \times 7^1, m\in \mathbb{N}$, and tested whether these numbers satisfy also the other bases representations.
While such numbers are likely to exist, our attempts for finding such a number have failed, and we conclude that no such number exists that is smaller than $10^{19}$. Furthermore, even if we found such a number, once we plug-in the number to the model, it might predict a value larger than 0, and even if our model predicts a value less than 0, it is very well likely that our model does not perform that well when considering numbers so much larger than those it was trained on.
\begin{table}[h!]
\centering
 \begin{tabular}{|c | c |} 
 \hline
 
  Base & 10 least significant digits \\
\hhline{|=|=|}
  Base 2  &  0, 0, 1, 0, 1, 0, 0, 0, 0, 0\\
 \hline
 Base 3 & 2, 0, 2, 0, 2, 1, 2, 0, 0, 2\\ 
 \hline
Base 5 & 0, 0, 0, 0, 0, 0, 0, 0, 1, 4  \\
 \hline
 Base 7 & 6, 1, 0, 0, 0, 0, 0, 6, 4, 0 \\
 \hline
\end{tabular}
\caption{Base representations of a hypothetical even number violating Goldbach's conjecture, with our model predicting a negative value of Goldbach's partitions for it.}
\label{tbl:magicalNum}
\end{table}

\subsection{Feature Analysis}
In this section we analyze the contribution each of the features has on the performance of the model. Table \ref{tbl:features} presents the performance of the model (in mean squared error) when trained without each of the following sets of features: base 2, 3, 4,  5 and 7 representation, without the number itself and without its log. As can be seen in the table, the base-3 features seem to have the greatest impact on the model, as removing them results with the highest error. Next in importance are the base-7 features. While, base-2 and base-5 seem to have a positive impact on the model, removing each of them separately does not harm the model's performance that much. 
Interestingly, the number itself turned out to be the least important feature. We also trained and evaluated the model while using only the single least significant digit of each of the bases; not surprisingly, this model did not perform well.

\begin{table}[h!]
\centering
 \begin{tabular}{|c | c |} 
 \hline
  Features used in model & MSE \\
\hhline{|=|=|}
 Without base 2  & 138,369\\
 \hline
 Without base 3 & 419,002\\ 
 \hline
Without base 5 & 112,653   \\
 \hline
 Without base 7 & 252,696  \\
 \hline
  Without log & 135,153 \\
 \hline
  Without number & 99,463  \\
 \hline
  Least significant digits & 391,707 \\
 \hline
  Full model (all features) & 89,764 \\ 
 \hline
\end{tabular}
\caption{Mean squared error (MSE) of model trained on a subset of the features.}
\label{tbl:features}
\end{table}

\section{Discussion} 
As stated in the introduction, Goldbach's conjecture has been verified up-to $4 \times 10^{18}$. This verification was performed by using exhaustive search. Our approximation model may allow a selective search method in which Goldbach's conjecture can be verified only for suspicious numbers according to our model, that is, only numbers that our model predicts will have a very low number of pairs. This approach can also be used to find numbers that violate the lower-bound proposed by \cite{provatidis2013rule}.
However, such selective search may require retraining our model on data closer to the target distribution (i.e., larger numbers), and adding additional digits to the base representations. 

The success of our method can be attributed, for the most part, to the base representations added as features. In our work we used based representations for the first 4 prime numbers (2, 3, 5, 7), though it is likely that adding few additional base representations with the following prime numbers (e.g. 11, 13, 17), would increase the model's accuracy. However, it is impractical to add more than a few additional representations (adding all prime representations up to the given number  would require prime factorization, which is the exact problem our method tries to avoid).  

While deep learning has shown great success in many different fields \cite{lv2015traffic,cruz2013deep,alipanahi2015predicting}, we believe that the success shown in this paper related to an open mathematical problem in number theory, is a big step and should not be disregarded as being merely another deep learning application. Our work may lead to a new paradigm of using deep learning (or machine learning in general) to solve mathematical problems such as prime factorization, friendly numbers, finding prime twins and many similar problems, which may currently seem out of the scope of deep learning methods. 

\section{Conclusions}

Goldbach's conjecture and Goldbach's function have remained open mathematical questions for over two and a half centuries. There have been several analytic attempts to approximate Goldbach's function, but unfortunately, these approximations either do not work well in practice or require prime factorization (prime decomposition) which is a hard problem.
In this paper we present the first deep-learning based approach to approximating Goldbach's function. We show that our approach outperforms current state-of-the-art approximations while not requiring prime factorization. We believe that our results can bring us one step closer to solving one of the worlds most significant open mathematical question.




\bibliographystyle{named}
\bibliography{goldbach}

\begin{thebibliography}{}

\bibitem[\protect\citeauthoryear{Alipanahi \bgroup \em et al.\egroup
  }{2015}]{alipanahi2015predicting}
Babak Alipanahi, Andrew Delong, Matthew~T Weirauch, and Brendan~J Frey.
\newblock Predicting the sequence specificities of dna-and rna-binding proteins
  by deep learning.
\newblock {\em Nature biotechnology}, 33(8):831, 2015.

\bibitem[\protect\citeauthoryear{Baker}{2007}]{baker2007excel}
John Baker.
\newblock Excel and the goldbach comet.
\newblock {\em Spreadsheets in Education (eJSiE)}, 2(2):2, 2007.

\bibitem[\protect\citeauthoryear{Bibel}{2013}]{bibel2013automated}
Wolfgang Bibel.
\newblock {\em Automated theorem proving}.
\newblock Springer Science \& Business Media, 2013.

\bibitem[\protect\citeauthoryear{Buhler \bgroup \em et al.\egroup
  }{1993}]{buhler1993factoring}
Joe~P Buhler, Hendrik~W Lenstra, and Carl Pomerance.
\newblock Factoring integers with the number field sieve.
\newblock In {\em The development of the number field sieve}, pages 50--94.
  Springer, 1993.

\bibitem[\protect\citeauthoryear{Colton \bgroup \em et al.\egroup
  }{2000}]{colton2000automatic}
Simon Colton, Alan Bundy, and Toby Walsh.
\newblock Automatic identification of mathematical concepts.
\newblock In {\em ICML}, pages 183--190, 2000.

\bibitem[\protect\citeauthoryear{Cruz-Roa \bgroup \em et al.\egroup
  }{2013}]{cruz2013deep}
Angel~Alfonso Cruz-Roa, John Edison~Arevalo Ovalle, Anant Madabhushi, and Fabio
  Augusto~Gonz{\'a}lez Osorio.
\newblock A deep learning architecture for image representation, visual
  interpretability and automated basal-cell carcinoma cancer detection.
\newblock In {\em International Conference on Medical Image Computing and
  Computer-Assisted Intervention}, pages 403--410. Springer, 2013.

\bibitem[\protect\citeauthoryear{Fliegel and
  Robertson}{1989}]{fliegel1989goldbach}
Henry~F Fliegel and Douglas~S Robertson.
\newblock Goldbach's comet: the numbers related to goldbach's conjecture.
\newblock {\em Journal of Recreational Mathematics}, 21(1):1--7, 1989.

\bibitem[\protect\citeauthoryear{Goldbach}{1742}]{goldbach7letter}
Christian Goldbach.
\newblock Letter to l.
\newblock {\em Euler, June}, 7, 1742.

\bibitem[\protect\citeauthoryear{Hardy and Littlewood}{1922}]{hardy1922some}
Godfrey~H Hardy and John~E Littlewood.
\newblock Some problems of diophantine approximation: The lattice-points of a
  right-angled triangle.
\newblock {\em Proceedings of the London Mathematical Society}, 2(1):15--36,
  1922.

\bibitem[\protect\citeauthoryear{Helfgott}{2013}]{helfgott2013ternary}
Harald~A Helfgott.
\newblock The ternary goldbach conjecture is true.
\newblock {\em arXiv preprint arXiv:1312.7748}, 2013.

\bibitem[\protect\citeauthoryear{Kingma and Ba}{2014}]{kingma2014adam}
Diederik~P Kingma and Jimmy Ba.
\newblock Adam: A method for stochastic optimization.
\newblock {\em arXiv preprint arXiv:1412.6980}, 2014.

\bibitem[\protect\citeauthoryear{Larson}{2005}]{larson2005survey}
Craig~E Larson.
\newblock A survey of research in automated mathematical conjecture-making.
\newblock {\em DIMACS Series in Discrete Mathematics and Theoretical Computer
  Science}, 69:297, 2005.

\bibitem[\protect\citeauthoryear{Lenat and Brown}{1984}]{lenat1984and}
Douglas~B Lenat and John~Seely Brown.
\newblock Why am and eurisko appear to work.
\newblock {\em Artificial intelligence}, 23(3):269--294, 1984.

\bibitem[\protect\citeauthoryear{Lenat}{1977}]{lenat1977automated}
Douglas~B Lenat.
\newblock Automated theory formation in mathematics.
\newblock In {\em Proceedings of the 5th international joint conference on
  Artificial intelligence-Volume 2}, pages 833--842. Morgan Kaufmann Publishers
  Inc., 1977.

\bibitem[\protect\citeauthoryear{Lenat}{1983}]{lenat1983eurisko}
Douglas~B Lenat.
\newblock Eurisko: a program that learns new heuristics and domain concepts:
  the nature of heuristics iii: program design and results.
\newblock {\em Artificial intelligence}, 21(1-2):61--98, 1983.

\bibitem[\protect\citeauthoryear{Lv \bgroup \em et al.\egroup
  }{2015}]{lv2015traffic}
Yisheng Lv, Yanjie Duan, Wenwen Kang, Zhengxi Li, and Fei-Yue Wang.
\newblock Traffic flow prediction with big data: a deep learning approach.
\newblock {\em IEEE Transactions on Intelligent Transportation Systems},
  16(2):865--873, 2015.

\bibitem[\protect\citeauthoryear{Markakis \bgroup \em et al.\egroup
  }{2012}]{markakis2012some}
Emmanuel Markakis, Christopher Provatidis, and Nikiforos Markakis.
\newblock Some issues on goldbach conjecture.
\newblock {\em Number Theory}, 29, 2012.

\bibitem[\protect\citeauthoryear{Montgomery and
  Vaughan}{1975}]{montgomery1975exceptional}
H~Montgomery and R~Vaughan.
\newblock The exceptional set of goldbach's problem.
\newblock {\em Acta Arithmetica}, 27(1):353--370, 1975.

\bibitem[\protect\citeauthoryear{Newell and Simon}{1956}]{newell1956logic}
Allen Newell and Herbert Simon.
\newblock The logic theory machine--a complex information processing system.
\newblock {\em IRE Transactions on information theory}, 2(3):61--79, 1956.

\bibitem[\protect\citeauthoryear{Oliveira~e Silva \bgroup \em et al.\egroup
  }{2014}]{oliveira2014empirical}
Tom{\'a}s Oliveira~e Silva, Siegfried Herzog, and Silvio Pardi.
\newblock Empirical verification of the even goldbach conjecture and
  computation of prime gaps up to 4⋅ 10$^1$⁸.
\newblock {\em Mathematics of Computation}, 83(288):2033--2060, 2014.

\bibitem[\protect\citeauthoryear{Prechelt}{1998}]{prechelt1998automatic}
Lutz Prechelt.
\newblock Automatic early stopping using cross validation: quantifying the
  criteria.
\newblock {\em Neural Networks}, 11(4):761--767, 1998.

\bibitem[\protect\citeauthoryear{Provatidis \bgroup \em et al.\egroup
  }{2013}]{provatidis2013rule}
Christopher Provatidis, Emmanuel Markakis, and Nikiforos Markakis.
\newblock Rule of thumb bounds in goldbach’s conjecture.
\newblock {\em American Journal of Mathematical Analysis}, 1(1):8--13, 2013.

\bibitem[\protect\citeauthoryear{Wang}{2002}]{wang2002goldbach}
Yuan Wang.
\newblock {\em The Goldbach Conjecture}, volume~4.
\newblock World scientific, 2002.

\bibitem[\protect\citeauthoryear{Whitehead and
  Russell}{1912}]{whitehead1912principia}
Alfred~North Whitehead and Bertrand Russell.
\newblock {\em Principia mathematica}, volume~2.
\newblock University Press, 1912.

\end{thebibliography}
\end{document}